\documentclass[letterpaper]{article} % DO NOT CHANGE THIS
\usepackage{aaai2026}  % DO NOT CHANGE THIS
\usepackage{times}  % DO NOT CHANGE THIS
\usepackage{helvet}  % DO NOT CHANGE THIS
\usepackage{courier}  % DO NOT CHANGE THIS
\usepackage[hyphens]{url}  % DO NOT CHANGE THIS
\usepackage{graphicx} % DO NOT CHANGE THIS
\usepackage{natbib}  % DO NOT CHANGE THIS AND DO NOT ADD ANY OPTIONS TO IT
\usepackage{caption} % DO NOT CHANGE THIS AND DO NOT ADD ANY OPTIONS TO IT
\usepackage{algorithm}
\usepackage{algorithmic}

\usepackage{newfloat}
\usepackage{listings}
\DeclareCaptionStyle{ruled}{labelfont=normalfont,labelsep=colon,strut=off} % DO NOT CHANGE THIS
\floatstyle{ruled}
\newfloat{listing}{tb}{lst}{}
\floatname{listing}{Listing}
\usepackage{amsmath}    % For \text, \eqref, and math environments
\usepackage{amssymb}    % For math symbols like \mathbb
\usepackage{amsfonts}   % For \mathbb{N}, etc.
\usepackage{amsthm}
\usepackage{mathtools}  % Optional but helpful extensions to amsmath
\usepackage{enumitem}   % For custom itemize formatting
\usepackage{xcolor}
\usepackage{placeins}
\usepackage{pifont}

\newcommand{\cmark}{\ding{51}}  % ✓
\newcommand{\xmark}{\ding{55}}  % ✗

\newtheorem{theorem}{Theorem}

\newtheorem{lemma}{Lemma}[theorem]

\title{Fast Task Adaptation in Meta Reinforcement Learning via \\ Latent Hypothesis-Planned Exploration}
\author{
    Maxwell J. Jacobson,
    Rohan Menon,
    John Zeng,
    Yexiang Xue
}
\affiliations{
    Purdue University\\
    \{jacobs57,menon67,zeng261,yexiang\}@purdue.edu
}

\usepackage{bibentry}
\begin{document}

\maketitle

\begin{abstract}
Meta-Reinforcement Learning (Meta-RL) learns optimal policies across a series of related tasks. A central challenge in Meta-RL is rapidly identifying which previously learned task is most similar to a new one, in order to adapt to it quickly. Prior approaches, despite significant success, typically rely on passive exploration strategies such as periods of random action to characterize the new task in relation to the learned ones. While sufficient when tasks are clearly distinguishable, passive exploration limits adaptation speed when informative transitions are rare or revealed only by specific behaviors. We introduce Hypothesis-Planned Exploration (HyPE), a method that actively plans sequences of actions during adaptation to efficiently identify the most similar previously learned task. HyPE operates within a joint latent space, where state-action transitions from different tasks form distinct paths. This latent-space planning approach enables HyPE to serve as a drop-in improvement for most model-based Meta-RL algorithms. By using planned exploration, HyPE achieves exponentially lower failure probability compared to passive strategies when informative transitions are sparse. On a natural language Alchemy game, HyPE identified the closest task in 65–75\% of trials, far outperforming the 18–28\% passive exploration baseline, and yielding up to 4× more successful adaptations under the same sample budget.
\end{abstract}

\begin{figure*}[t]
    \centering
    \includegraphics[width=0.80\linewidth]{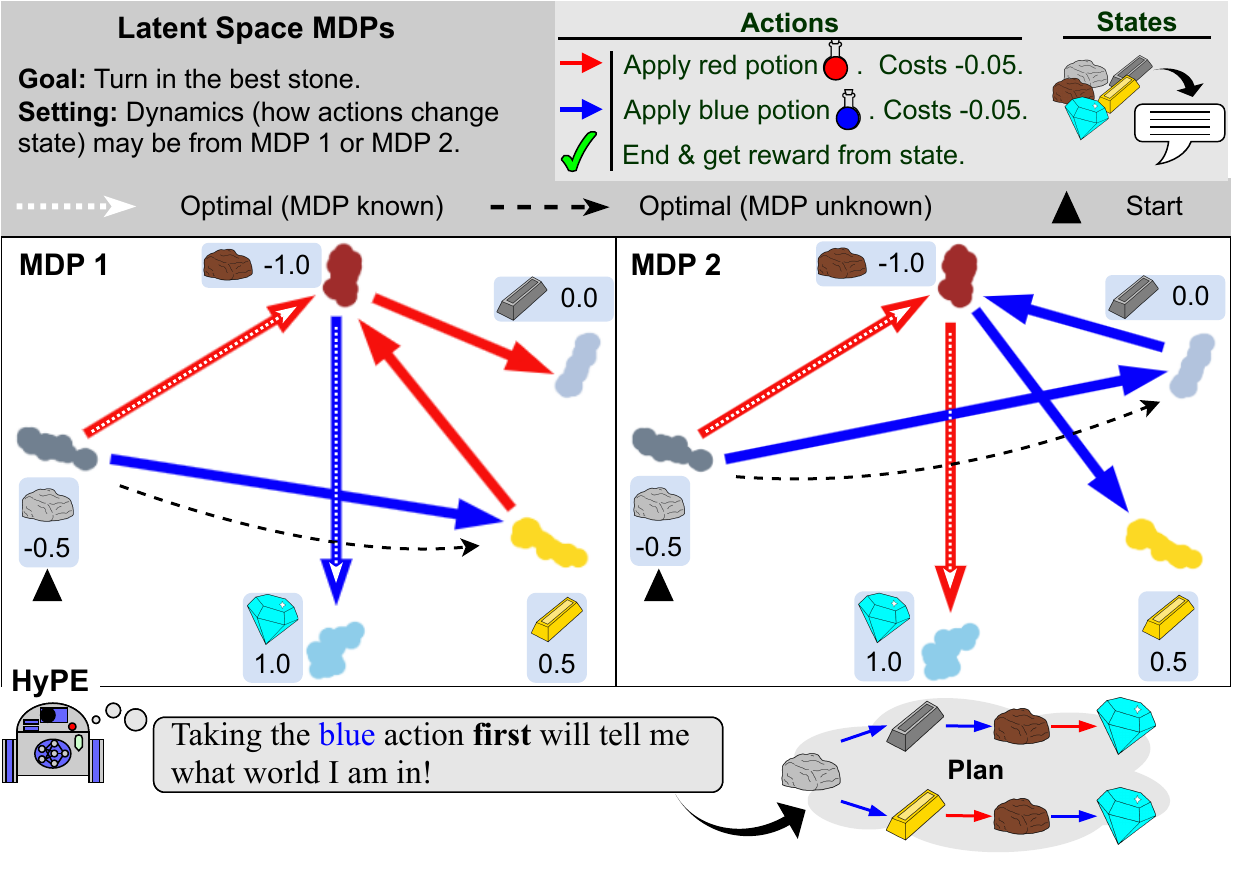}
    \vspace{-10pt}
    \caption{Latent space dynamics of a natural language Alchemy game, where the goal is to maximize the value of stones before \textit{turning in} and ending the game. Stone states, such as "a sample of diamond" or "a bar of gold," are conveyed to the agent in natural language and transformed into a latent representation visualized here. The agent begins with a gray stone, aiming to reach the optimal end state: diamond. Two MDPs represent the chemical behavior on potions transform stones (colored arrows). In both MDPs, the shortest path to diamond starts with the red action. However, the agent does not initially know which MDP governs the game, making red a suboptimal first choice due to ambiguity about subsequent actions. Instead, starting with the blue action is more effective, as it reveals the active MDP: in MDP 1, blue leads to gold, while in MDP 2, it leads to iron. HyPE operates in the latent space to identify trajectories that distinguish between the two MDPs (e.g., the blue action diverges to gold in one and iron in the other). Once the active MDP is identified, the path to diamond becomes straightforward, allowing the agent to plan optimally. In this demonstration, HyPE is able to discover the true MDP on the first transition 100\% of the time.} 
    \label{fig:main_fig}
\end{figure*}

\section{Introduction}
Reinforcement Learning -- despite its ability to learn in diverse environments -- remains slow to adapt when underlying dynamics change rapidly, posing an obstacle in many real-world applications. If an agent cannot align itself with new situations -- new \textit{tasks} -- quickly, it may fail to learn the task at all before further changes emerge, making sample efficiency a paramount concern.

Meta-Reinforcement Learning (Meta-RL) addresses this shortcoming by training agents on a variety of tasks in a meta-training phase, enabling them to quickly adapt to new, unseen tasks in an adaptation phase with minimal additional training samples. Some of the most effective approaches for improving sample efficiency are model-based meta-RL methods. They learn one model for each task during meta-training. These models predict how actions taken in a given state will affect the rewards and the next state, which may vary across tasks. However, during the adaptation phase, determining which model is best suited to adapt from -- the most similar to the current task -- remains an open challenge.

A key limitation in these meta-model methods is their dependency on passive model exploration. 
By observing changes without actively probing them, agents struggle to rapidly grasp the \textit{transition dynamics} behind new tasks -- wasting adaptation samples on uninformative actions. 
This restricts the ability to infer the closest model for new tasks.

We introduce \textbf{Hy}pothesis-\textbf{P}lanned \textbf{E}xploration (HyPE), a method that employs hypothesis forming and verification to shift to active identification of the correct closest transition model. HyPE collects a pool of candidate dynamics models and selects actions which most efficiently reduce uncertainty about which model best fits the environment. By treating adaptation as a scientific experiment to confirm or disprove hypotheses about the environment, HyPE maximizes useful information gained, enabling efficient selection of the model closest to the current task for further adaptation. %, which can then be .

HyPE is also designed for \textit{planning \textbf{within} the latent space} of deep dynamics models. This means that it can be used as a drop-in improvement to a wide family of meta-RL algorithms regardless of underlying transition dynamics, problem structure, or modality. This also means that HyPE does not rely on any model being completely correct or any single transition providing a clear indicator -- it finds the model with the most similar latent dynamics, even if some transitions are different. This planned exploration strategy fills the gap left by passive approaches by selecting the closest model from the meta-trained pool using fewer samples, enabling faster adaptation and better performance under rapidly changing conditions.

As a motivating example, consider the natural language alchemy game, which we will use as one experiment domain. A sample game is shown in Figure~\ref{fig:main_fig}. In this game, the agent must use potions to change the states of materials, turning them into higher-value ones (e.g., turning stone into gold). Stone states are communicated to the agent in natural language, such as ``a bar of gold''. The agent starts at a gray stone, and the optimal end state is a diamond. There are two possible ``chemistry'' (i.e., tasks) -- represented as Markov Decision Processes (MDPs) -- which define how potions change each stone type (e.g., does applying the blue potion to gray stone turn it into iron or gold?). In both chemistries, the quickest path to the diamond (hence with the highest reward) begins with the action of using the red potion.

Will you use the red potion in the first step, if you do not know which chemistry the current world operates?
Notice that the unknown tasks (i.e., chemistry in the game) are a characteristic of meta-RL. 
Interestingly, the optimal move is to use the blue potion first, 
because taking this action and examining whether the next state is iron or gold reveals the correct chemistry. 
You can view the blue action as a \textit{scientific experiment}, which rules out implausible worlds. 
HyPE incorporates this strategy by planning an action sequence that maximizes the separation between the learned MDPs. 
Then, HyPE can choose an MDP model in its learned pool that best fits the outcomes of the planned sequence, allowing us to quickly adapt to the correct MDP during test time. 
Note that the chosen model need not exactly match the true MDP at test time because the model can still adapt as it explores. It just needs to be fairly close so that adaptation is possible with only a few additional data points.

HyPE demonstrated superior adaptation efficiency in complex natural language Alchemy games by correctly identifying the closest model in 65-75\% of trials, compared to 18-28\% for a passive exploration baseline. This higher accuracy led to faster convergence and significantly higher rewards. HyPE successfully solved the Alchemy game within the allotted adaptation time up to four times more frequently compared to the baseline.

\section{Preliminaries}

\subsection{Reinforcement Learning (RL)}

Reinforcement Learning (RL) is a machine learning paradigm in which an agent interacts with an environment to maximize a reward signal. The agent learns by choosing actions and observing the resulting outcomes and transitions, aiming to develop an optimal policy.

This process is often framed as a Markov Decision Process (MDP) $\langle S, A, P, R, \delta \rangle$, defined by a \textit{state space} ($S$), which represents all possible configurations of the environment, an \textit{action space} ($A$), denoting the possible actions that change the state, a \textit{reward function} ($R: S \times A \rightarrow \mathbb{R}$), quantifying the immediate benefit or cost of a state-action pair, and the \textit{transition dynamics} ($P: S \times A \times S \rightarrow \mathbb{R}_+$), which describe the probability of moving from one state to another given an action. Additionally, \textit{discount factor} ($\delta \in [0, 1)$) is often included to reflect preference for immediate rewards over future ones. At each step, the agent occupies a state, selects an action, and observes the new state resulting from the transition dynamics, along with a reward signal.

The agent, tasked with optimizing the long-term return ($\sum_{i=0}^\infty \delta^i r_{i}$) of the MDP, learns some policy for choosing actions given a state ($\pi_\theta: S \rightarrow A$). When the agent takes a single step in the environment, the tuple of initial state, chosen action, received reward, and next state $(s_t, a_t, r_t, s_{t+1})$ is called a \textit{transition}. A sequence of transitions, representing a path through the MDP, is a trajectory (e.g. $\tau = \{(s_0, a_0, r_0, s_1), (s_1, a_1, r_1, s_2), \dots\}$). The agent must also balance exploring unknown dynamics with exploiting known strategies to optimize long-term reward.

\subsection{Meta-RL Framework (meta-RL)}

Meta-RL is a field that seeks to apply meta-learning \cite{Vilalta2002,huisman2021survey,WANG202190} to reinforcement learning, resulting in a slightly changed problem structure. In this setting, we hope to learn a set or distribution of MDPs ($M_i \sim \mathcal{M}$). Generally, we consider some aspects of the MDPs to remain consistent across the set: The state space, action space, and discount factor are usually equal across all. The reward function and transition dynamics are subject to change. Meta-RL MDPs are then formulated as $M_i = \langle S, A, P_i, R_i, \delta \rangle$.

This changes the goal of meta-RL as well. Instead of optimizing a single policy for a fixed MDP, the aim is to develop a meta-policy $\pi_\phi$ capable of maximizing reward across any MDP in the set $\mathcal{M}$, given a limited period for adaptation ($\pi_\phi \rightarrow \pi_i$). It follows, then, that meta-RL is typically divided into two phases: \textit{meta-training}, where the meta-policy $\pi_\phi$ is trained across a diverse set of MDPs from $\mathcal{M}$, and \textit{adaptation} (also known as \textit{meta-testing}), where $\pi_\phi$ adapts to a specific MDP $M_i \in \mathcal{M}$ using limited interaction data, resulting in the adapted policy $\pi_i$ which maximizes the long term returns of $M_i$ specifically.

In other words, most meta-RL has a large number of MDPs sampled from $\mathcal{M}$ during meta-training, where they are used in learning $\phi$. During this phase, the MDPs are generally known -- the agent can know which MDP of the set it is currently acting on. In the adaptation phase, a new, potentially unseen MDP is sampled from $\mathcal{M}$. It might be very similar to a previously-seen example, or very different. The goal is always to optimize for this new MDP given few samples, leveraging prior knowledge from meta-training.

RL and meta-RL methods are categorized into two primary approaches:
\textit{model-free RL}, which learns policies directly from interaction data without explicitly modeling transition dynamics, and \textit{model-based RL}, which builds a model of the transition dynamics to predict and plan future outcomes. \textbf{This work considers model-based meta-RL}. Under this setting, we learn a network to approximate $P_i$ and $R_i$ directly (model $\langle \hat{P_i}, \hat{R_i} \rangle \approx \langle P_i, R_i \rangle$), and then derive a policy $\pi_i$ from this model. This can be done with planning methods, such as \textit{model predictive control} \cite{camacho2013model} (MPC). If $\hat{P_i}$ and $\hat{R_i}$ are represented as a set of neural networks, then the basic process during meta-training is to gather transitions (often randomly), and to train each network to predict the next state and reward given the current state and action on its own indexed MDP. 

After meta-training there is a pool of models $\mathcal{H}$, with each model trained on its own task ($M_i$). When the adaption phase begins, these algorithms must choose one of the models: one with dynamics as similar as possible to the current unknown (perhaps unseen) task. Generally, these algorithms use \textbf{passive exploration} here. One example is the explore-then-commit (EtC) strategy: (1) take random actions for a short time ($k$ steps), (2) evaluate each model against the collected transitions, and (3) select the model with the closest predictions.

The model is selected by finding a $\langle\hat{P},\hat{R}\rangle \in \mathcal{H}$ that minimizes Equation~\ref{eq:select_model}.

\begin{equation}
\frac{1}{|B|} \sum_{(s, a, s') \in B} \| s' - \hat{P}(s, a) \|_2^2
\label{eq:select_model}
\end{equation}

This represents the mean squared error of any given model's predictions versus the true next state ($s'$), where buffer $B$ contains random-action transitions from passive exploration, and $\hat{P}(s, a)$ is the predicted next state of model $\hat{P}$. This adopted model can then be further adapted using online training, collecting transitions into a buffer, and then updating the latent dynamics model with them periodically.

\subsection{Latent Vectors as a General Representation}

To learn systems with complex inputs, a common deep learning approach is to leverage \textit{latent spaces}, which distill high-dimensional input data into low-dimensional, meaningful representations. Using autoencoders, text or image inputs can be compressed to a small vector including essential features in a learned representation.

This is especially useful in model-based meta-RL. Instead of taking raw state and action as input and outputting a prediction of the next state, latent dynamics models use the encoded state -- a latent state -- to predict a distribution (for example, mean and variance of a Gaussian) over the next latent state. In other words, we learn an encoder $Enc$ which maps elements of $S$ to a low dimensional vector. Then, the dynamics model can learn the transition dynamics over this encoded space instead of the higher-dimensional data ($\hat{P}_i(\mathrm{Enc}(s_{t+1}) \mid \mathrm{Enc}(s_t), a_t)$). In practice, the model samples next latents in stochastic cases ($z_{t+1} \sim \hat{P}_i(\cdot \mid z_t, a_t)$), or returns a point estimate in deterministic ones (e.g., $z_{t+1} = \arg\max_z \hat{P}_i(z \mid z_t, a_t)$). The lower dimensionality makes learning and planning easier in this space, assuming the encoder itself is well-trained. 

In meta-RL, the encoder and its latent space can generally be shared among all MDPs in $\mathcal{M}$. This is notable in Figure~\ref{fig:main_fig}. Here, the colored points in the space are the encoded latent representations of these (natural language in this case) states. Because the encoder there is well-trained, similar states (color) are grouped spatially close to each other, forming natural clusters. If the dots are latent states, the arrows then represent the latent dynamics. We can see how taking a particular action can move a dot from the vicinity of one cluster to the vicinity of another.

\section{Hypothesis-Planned Exploration}

Prior work in model-based meta-RL often relies on passive exploration to adapt to new dynamics, which can lead to inefficiency when multiple models in the pool share significant overlap \cite{kaushik2020fast}. These random trials waste resources that could instead refine the agent's understanding of critical transitions. Hypothesis-Planned Exploration (HyPE) addresses this inefficiency by actively planning experiments to distinguish between competing dynamics models, thereby accelerating adaptation and improving sample efficiency.

\subsection{Within the Meta-RL Framework}

\begin{figure*}[t]
\centering
\includegraphics[width=0.9\linewidth]{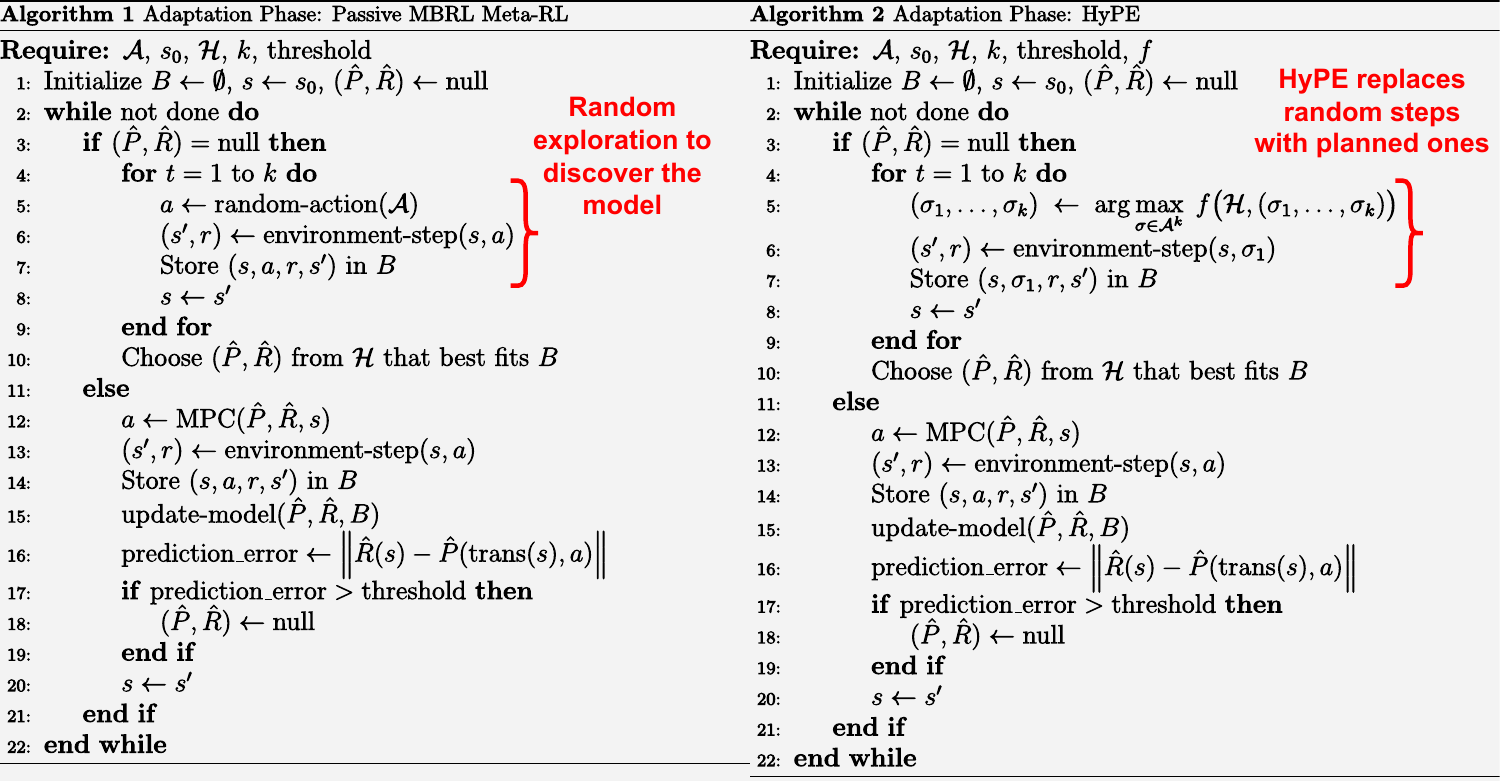}

\caption{Comparison of typical adaptation algorithm used in passive exploration meta-RL (left) versus HyPE (right). Roughly this procedure is used by several works, any of which can be upgraded to use HyPE, because it only requires replaces the step of choosing which model from the pool to build on (inside \texttt{if $\langle \hat{P}_i, \hat{R}_i \rangle = null$}). Instead of spending $k$ steps exploring randomly (in this case, explore-then-commit strategy), these steps are planned to maximize seperation between the models.}%\cite{kaushik2020fast}
\label{fig:algo}
\end{figure*}

As mentioned, meta-RL is usually divided into a meta-training phase (example tasks from a set $M_0, M_1, M_2, ... \sim \mathcal{M}$ are learned by models) and an adaptation phase (the new task is learned, with information from meta-training as a prior). HyPE's role is to choose a closest model from a pool during the adaptation phase -- doing so more efficiently and accurately than passive methods.

HyPE considers the set of models learned in meta-training $\mathcal{H}$ to be \textit{hypothesis models}, each representing a distinct set of transition dynamics $\hat{P}_i \approx P_i$ and reward function $\hat{R}_i \approx R_i$ within a shared latent space defined by an encoder $Enc$. HyPE discovers the current task by planning a sequence of actions (we call this \textit{experimental action sequence} $\sigma$) that maximizes separation between models according to a \textit{separating function}. These experiments are designed to maximize the information gained in relation to $\mathcal{H}$, enabling rapid convergence to the most accurate model.

\subsection{Planned Experimentation for Model Selection}
The HyPE process begins by defining a maximum number of steps, $k$, for the planned experiment. The agent optimizes over possible action sequences to find the one that maximizes separation between models in the pool: $\arg\max_{\sigma \in \mathcal{A}^k} f_s(\mathcal{H}, \sigma, s_0)$. Here, we are finding the $k$-length action sequence $\sigma$ which maximizes the separating function $f_s$ from the initial state. The separating function takes as input the pool of previously learned models, the candidate action sequence, and the state.

This paper offers several separating functions for different scenarios (see Appendix A), which measure how effectively an action sequence distinguishes candidate dynamics models. For example, Figure~\ref{fig:main_fig} shows a scenario with two latent-space MDPs, each having different transition dynamics. If we already know the correct MDP, the optimal action is applying the red potion first. But when considering uncertainty in environment dynamics, it is better to first apply the blue potion, since that would lead to clearly different latent states -- silver or gold -- depending on the true MDP, allowing us to quickly identify it and ultimately reach the diamond. Intuitively, the red action yields little separation, as the resulting states are very close to each other in the latent space (both in brown stone cluster) in each MDP. Whereas, the blue action yields large separation (the distance between the silver and gold clusters).

The separating function depicted in the example is called L2-Across, which sums latent-space L2 distances between every pair of models' predicted paths. Because this scales quadratically with model count, we use the faster approximation Central Deviation (CD), which instead measures the distance of each model's predictions to the average prediction, significantly reducing computation. Note that both of these are specialized for deterministic MDPs, assuming that hypothesis models predict a single next latent point rather than a distribution.

To handle the general stochastic case, we define analogous functions based on KL divergence. Specifically, Pairwise KL (PKL) measures pairwise distributional disagreements among models, while Central KL Divergence (CKLD) approximates this efficiently by measuring each model’s divergence from the average prediction. Appendix A offers the formal specification of all separating functions, and discussions on when each is applicable.

Once the experimental sequence $\sigma$ is executed and transitions are collected in the experience buffer ($B$), HyPE adopts the hypothesis with dynamics most plausibly generating those samples. This can be done with MSE if the hypothesis pool uses deterministic models, where MSE measures the average squared discrepancy between predicted and observed latents for a fast, direct fit metric. Alternatively, for stochastic models that output a mean and variance, we use negative log-likelihood, which scores each transition by its probability under the model’s distribution, intuitively accounting for both prediction error and model confidence.

The key to HyPE is then the informational quality of the transitions in $B$. In passive exploration, the majority of $B$ may be filled with transitions that yield no useful information -- where the outcome of a state-action pair is roughly the same next-state in every model. In HyPE, planning ensures that we fill the buffer with as many informative transitions as possible, resulting in a much better prediction of $\langle\hat{P},\hat{R}\rangle$ from $\mathcal{H}$ given the same number of exploration samples.

The generality of HyPE can be seen in Figure~\ref{fig:algo}, requiring no changes to the meta-training procedure of a meta-RL algorithm, and utilizing the same adaptation strategy after model selection of any compliant method as well. By utilizing the existing pool of models to do a small amount of additional planning, the sample efficiency of the adaptation phase can be drastically improved on many environments.

If the selected model's accuracy degrades during subsequent transitions (e.g., a drop in latent state prediction accuracy below a threshold), the agent can un-adopt the model and repeat the HyPE process to identify a better-fitting model.

\section{Theoretical Analysis}

In many real environments, only a subset of transitions differentiates competing dynamics. Random exploration (e.g., explore-then-commit in meta-RL) rarely encounters these when they are hard to reach. HyPE uses planned exploration to target them directly, increasing the chance of correct dynamics selection. Here, we formalize this advantage.

HyPE reduces sample needs by an \textit{informative occupancy ratio} (IOR) and achieves exponentially lower failure probability than uniform strategies under fixed exploration budgets when informative transitions are rare.

Consider $m$ hypothesis models $H_1,\dots,H_m \in \mathcal{H}$ sharing finite state space $\mathcal S$, action space $\mathcal A$, and initial distribution $\rho$. Each models an MDP that is ergodic (any state reachable from any other in finite time) and distinguishable (at least one differing transition). Let $\pi_{uni}(a \mid s) = 1/|\mathcal A|$ be a uniform random policy and $\pi_{hype}$ a planned policy, both run for $T$ steps to collect transitions $(s_t, a_t, s_{t+1})^{1,..,T}$. From this data, we form a maximum-likelihood estimate $\hat i$ of the true index $i^*$. $\pi_{hype}$ chooses actions $\sigma_{0,...,T}$ to maximize mutual information with the random variable $W$ realized as $i^*$.

To understand the gap, define the \textit{informative region} $\mathcal G \subseteq \mathcal S \times \mathcal A$ as state-action pairs with meaningfully differing next-state distributions (KL divergence greater than some $d_0$). Let $\varepsilon$ be the frequency with which $\pi_{uni}$ visits $\mathcal G$ (typically small), and $\alpha$ the frequency with which $\pi_{hype}$ visits $\mathcal G$ (typically larger). Spending actions outside $\mathcal G$ can only reveal useful distinguishing information very slowly. The IOR is defined as $\alpha/\varepsilon$, measuring how much more effectively $\pi_{hype}$ targets informative transitions. Large IOR implies reduced sample requirements and chance of error ($\hat i \neq i^*$).

\begin{theorem}
\label{thm:ior}

Uniform exploration accrues information at rate $\varepsilon d_0$ per step, while $\pi_{\text{hype}}$ accrues $\alpha d_0$ per step. Consequently, to achieve error $\delta$, planned exploration needs at most a $\frac{1}{\mathrm{IOR}}$ fraction of the samples. Moreover, for any fixed horizon $T$, the error ($\hat i \neq i^*$) ratio satisfies
\[
\frac{P_{\pi_{\text{hype}}}\bigl(\text{error}\bigr)}
{P_{\pi_{\text{uni}}}\bigl(\text{error}\bigr)}
\;\le\;
\exp\!\bigl(-(\alpha - \varepsilon)\,d_0\,T\bigr)
\]

an exponential separation in $T$ whenever $\alpha > \varepsilon$. 
\textit{In other words}, HyPE’s error probability is exponentially smaller:
\[
P_{\pi_{\text{hype}}}(\text{error})
\le
P_{\pi_{\text{uni}}}(\text{error}) \cdot \exp\!\bigl(-(\alpha - \varepsilon)\,d_0\,T\bigr)
\]
\end{theorem}

Consider an example environment with 100 states arranged linearly and cyclically from 1 to 100. The agent starts at a uniformly random state. At each state, the agent can take actions ``left'' or ``right.'' The ``left'' action deterministically moves to the previous state. The ``right'' action moves forward with $\sim 70\%$ chance and otherwise stays put -- except at state 50, where the MDPs differ. In MDP 1, ``right'' at state 50 succeeds with only 10\%, while in MDP 2 it succeeds with 90\%. Thus, the informative region $\mathcal G$ should consist only of (50, ``right''), and the divergence between these is $d_0 \approx 1.8$. All other transitions differ trivially (e.g., 70\%-30\% vs. 69\%-31\%, giving maximum divergence $\bar{d} \approx 0.00024$). A uniform random policy samples each action equally and visits all state-action pairs nearly uniformly, giving $\varepsilon \approx 0.005$. By contrast, HyPE directs the agent efficiently to state 50 and then alternates between (50, ``right'') and (49, ``left''), spending 1 out of every 2 steps in $\mathcal{G}$ thereafter; including the upfront cost to reach state 50, this yields $\alpha \approx 0.4$ by $T = 100$. In this environment $\varepsilon \ll \alpha$.

Theorem \ref{thm:ior} predicts an IOR of 80 and bounds the error ratio between policies by $\exp\bigl(-0.395 \times 1.8 \times 100\bigr) \approx 2.4 \times 10^{-31}$. That is, HyPE’s error probability is provably at least $10^{31}$ times smaller in this setting than $\pi_{uni}$ for fixed $T = 100$. This is reasonable because the chance of $\pi_{uni}$ ever seeing $\mathcal G$ is very unlikely over even 100 steps, and transitions outside of $\mathcal G$ offer such little information. In contrast, $\pi_{hype}$ can reach it and sample it many times over during the same period.

A full proof of Theorem~\ref{thm:ior} is provided in Appendix D. Is also proves a related result (Theorem~\ref{thm:closest}) showing that even when the true environment is not in the hypothesis pool, HyPE still enjoys an exponential advantage in identifying the closest matching MDP under KL divergence (as is the case in most meta-RL settings).

\begin{figure}[t]
    \centering
    \includegraphics[width=0.7\linewidth]{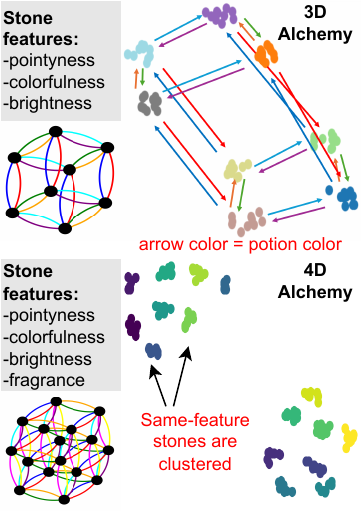}
    \caption{Latent dynamics of the natural language Alchemy environment with 3 and 4 binary features, as learned in the experiments. For each, the textual names of the features are in the top left gray box. These are encoded into the learned latent space to the right. Points are then latent-space representations of text like ``the stone is smooth, monochromatic, and bright''. Dynamics models learn how actions transition between these states (the arrows), and HyPE plans a sequence of actions to discover which dynamics model is most correct from a pool. The underlying dynamics model is plotted in the bottom left of each as a directed graph.}
    \label{fig:cube}
\end{figure}

\section{Related Work}
\subsection{Meta-Reinforcement Learning}

Meta-RL enables agents to generalize across diverse tasks through a meta-training phase and rapidly adapt to novel tasks during adaptation. Gradient-based approaches, such as MAML \cite{finn2017model,nichol2018first}, adapt tasks via gradient updates. Following work has refined gradient-based meta-training in a variety of ways \cite{li2017metasgd,sung2017learning,zintgraf2019fast,vuorio2019multimodal,song2020esmaml}. Recurrence-based approaches, including $RL^2$ \cite{duan2016rl,wang2016learning}, encode task histories for adaptation. Extensions have improved their handling of long-horizon tasks, task uncertainty, and incorporated modern attention mechanisms \cite{bhatia2023rl,zintgraf2020varibad,humplik2019meta,mishra2018simple,pmlr-v162-melo22a}. While these algorithms primarily focus on improving the meta-training phase, HyPE enhances the adaptation phase and is compatible with either family, serving as a drop-in improvement to accelerate adaptation.

\subsection{Model-based Meta-Reinforcement Learning}

Model-based meta-RL leverages predictive models of environment dynamics to enable efficient adaptation across tasks. During meta-training, agents learn shared representations of environment dynamics, which serve as a foundation for adaptation. Single-model approaches, such as CAMeLiD \cite{harrison2018control}, are well-suited to environments with slight variations, like quadrotor operation under varying payloads \cite{o2021meta}, but struggle to capture large or abrupt differences between tasks. Multi-model approaches, by contrast, handle diverse dynamics by maintaining a pool of models or task embeddings. It is this family that HyPE is clearly compatible with.

During adaptation, these methods typically refine models by either fine-tuning network parameters \cite{nagabandi2018learning,o2021meta,harrison2018control} or adjusting task embeddings that encode latent variables representing system-specific dynamics \cite{perez2018efficient,saemundsson2018meta,belkhale2021model}. However, selecting the correct model or embedding is also required. This is critical but often relies on passive exploration, where random actions guide model selection. FAMLE \cite{kaushik2020fast}, for example, is a multi-model method that takes a set number of random actions initially, observing the results, and choosing the model that most closely predicts each. This is also known as an explore-then-commit strategy.

\section{Experiments}

\subsection{Experimental Setup}

We evaluate HyPE using the natural language Alchemy environments, adapted from the Alchemy game~\cite{Wang2021AlchemyAB}. In these environments, the agent applies potions to transform a stone's binary attributes (e.g., sharp versus smooth, colorful versus dull), aiming to maximize stone value by reaching valuable states before turning in for a reward. The agent receives observations as natural language descriptions (e.g., ``The mineral is sharp, colorful, and dark''), requiring it to encode these observations into latent state representations using a pretrained DistilBERT-based encoder~\cite{Sanh2019DistilBERTAD}. We test two variants of increasing complexity: a 3D variant (8 underlying states) and a 4D variant (16 states). See Appendix B for more information on this environment.

Each meta-RL task is defined by unique transition dynamics (\textit{chemistry} of how potions change stones) and reward structures. Specifically, certain actions (potion applications) are blocked in some states for each task, altering paths through the latent state space. During meta-training, we sample a set of MDPs with distinct blocked transitions. In the adaptation phase, the agent encounters previously unseen MDPs created by introducing an additional blocked transition to existing meta-training MDPs, necessitating quick adaptation. Rewards vary per task based on attribute combinations, with additional penalties for longer paths. The structure of one of these chemistries is visualized both graphically and as latent states in Figure \ref{fig:cube}.

The core of our model-based meta-RL architecture is a pool of latent dynamics models that predict next latent states, rewards, and terminal conditions from current latent states and actions. When selecting optimal actions from the adopted hypothesis model, we employ Model Predictive Control (MPC) with random shooting to select actions based on the latent model predictions, planning several steps ahead. HyPE replaces the standard passive exploration step of the adaptation phase (such as in FAMLE's explore-then-commit method~\cite{kaushik2020fast}) with planned exploration that maximizes information gain about the current task's dynamics.

To isolate the effect of planned exploration, our baseline (EtC) uses the identical meta-training procedure and MPC-based adaptation policy but employs random action selection during initial exploration to choose the closest latent dynamics model from the meta-trained pool. Both methods share the exact same meta-training conditions, latent models, and adaptation algorithms, differing only in the initial model-selection exploration method.

Further experiment setup details, including text encoding, model architectures, reward specification, MDP sampling, computation resources, hyperparameters, and much more can be found in Appendix C.

\subsection{Experimental Results}

\noindent\textbf{HyPE chooses the correct closest model more often.} In the 3D setting with 8 underlying states, HyPE identified the closest model in 30 out of 40 trials, outperforming the EtC baseline, which identified the correct model in only 7 trials. In the 4D setting with 16 underlying states, HyPE correctly identified the closest model in 26 out of 40 trials, whereas the baseline identified the correct model in only 11 trials.

HyPE and EtC started with the same pool of models as a result of the meta-training phase. Testing each of these models on their own task, the original MDP they were trained on, yielded an average normalized reward across all tasks was $0.904 \pm 0.02$, with an average of $4 \pm 0.54$ steps required to reach the turn-in state in 3D. The 4D setting models yielded an average normalized reward of $0.891 \pm 0.01$ and an average of $4 \pm 0.56$ steps required. This confirms that each model was a fairly accurate approximation of its original MDP, but was far from a perfect one. This empirically shows that HyPE has some limited resilience to model error.

\begin{figure}[t]
    \centering
    \includegraphics[width=0.87\linewidth]{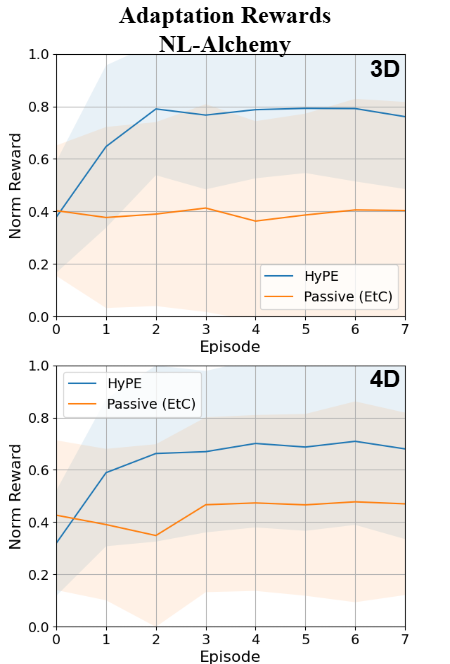}
    \vspace{-10pt}
    \caption{Classic Alchemy adaptation rewards for 3 feature (3D) and 4 feature versions (4D). HyPE converges much faster to the correct model, which translates to high reward in just a few episodes. Both methods use the same meta-trained model pool and adaptation procedure, and differ only on how they initially spend samples to select a model -- a passive exploration (EtC) versus a planned one (HyPE).}
    \label{fig:classic_reward_compare}
\end{figure}

\noindent\textbf{Correct closest model choice leads to faster adaptation.}  In the 3D setting, HyPE's ability to identify the correct model in 30 trials resulted in significant adaptation improvements. For example, all trials quickly surpassed a normalized reward of 0.2 in $1.05 \pm 0.32$ episodes, and 25 of the trials reached a reward above 0.8 in $4.64 \pm 1.93$ episodes. In contrast, the passive EtC approach only achieved a normalized reward above 0.8 in 6 trials. In the 4D setting, HyPE's correct identification in 26 trials also led to faster adaptation. All 40 trials achieved a normalized reward above 0.2 in $1.27 \pm 0.78$ adaptation episodes, and 13 trials reached a reward above 0.8 in $5.62 \pm 2.69$ episodes. By comparison, EtC achieved a normalized reward above 0.8 in only 9 trials. This is also shown clearly in Figure~\ref{fig:classic_reward_compare}. These plots show the mean reward across 40 trials over 8 adaptation episodes.

\section{Conclusion}
HyPE accelerates adaptation in model-based meta-RL by actively identifying the correct task dynamics model. Our experiments show that it consistently outperforms passive methods on natural language Alchemy tasks, achieving faster convergence and higher rewards. Beyond meta-RL model selection, HyPE’s hypothesis-planned exploration opens avenues for more comprehensive planning across pools of latent dynamics models in the latent space. Future work will explore this important area.

\FloatBarrier

\bibliography{main}

\clearpage
%\appendix
\onecolumn
\setcounter{section}{0}
\renewcommand{\thesection}{\Alph{section}}

\section{Appendix A: Separator Functions}
\label{sec:sepfun}

\FloatBarrier

\begin{table*}[h]
\centering
\label{tab:sepfunc_compare}
\begin{tabular}{lcccccc}
\hline
\textbf{Seperating Function} &
\rotatebox{90}{\shortstack{\textbf{Deterministic}}} &
\rotatebox{90}{\shortstack{\textbf{Stochastic}}} &
\rotatebox{90}{\shortstack{\textbf{Quadratic}\\\textbf{in \# Hypotheses}}} &
\rotatebox{90}{\shortstack{\textbf{Linear}\\\textbf{in \# Hypotheses}}} &
\rotatebox{90}{\shortstack{\textbf{Continuous}\\\textbf{State}}} &
\rotatebox{90}{\shortstack{\textbf{Discrete}\\\textbf{State}}} \\
\hline
Inconsistency (Incon)     & \cmark & \xmark & \cmark & \xmark & \cmark & \cmark \\
L2-Across (L2A)           & \cmark & \xmark & \cmark & \xmark & \cmark & \xmark \\
Central Deviation (CD)    & \cmark & \xmark & \xmark & \cmark & \cmark & \xmark \\
Pairwise KL (PKL)         & \xmark & \cmark & \cmark & \xmark & \cmark & \xmark \\
Central KL Divergence (CKLD) & \xmark & \cmark & \xmark & \cmark & \cmark & \xmark \\
\hline
\end{tabular}
\end{table*}

\subsection{Separating Functions for Deterministic MDPs}
For these, we assume the MDPs are deterministic, and so a neural model of $\hat{P}$  returns a single latent state prediction instead of an explicit distribution. In other words, $z_{t+1} = \text{infer}(s_t, a_t, \hat{P}) := \arg \max_z \hat{P}( z \mid Enc(s_{t}), a_t)$.

\noindent\textbf{Inconsistency (Incon).} A count of the pairwise separation between dynamics models. Here, $Enc$ is an encoder to transform states to a latent space. $\mathcal{H}$ is the set of models ($\langle \hat{P},\hat{R} \rangle \in \mathcal{H}$) learned during meta-training. $\sigma$ is a candidate sequence of actions, representing a possible experimental path. $s_0$ is the starting state from whence $\sigma$ starts, simulating each step forward with $\hat{P}_i$ and $\hat{P}_j$. $tol$ is a tolerance value, allowing this measurement to be meaningful in the continual context of latent spaces. If the distance between the two models' predictions is greater than tolerance, one is added to the count.

\begin{equation}
\text{Incon}(\mathcal{H}, \sigma, s_0) = \sum_{i=1}^{|\mathcal{H}|} \sum_{j=i+1}^{|\mathcal{H}|} \sum_{t=0}^{|\sigma|} \mathbf{1} \left( \| \text{infer}(s_t, \sigma_t, \hat{P}_i) - \text{infer}(s_t, \sigma_t, \hat{P}_j)  \| > \text{tol} \right)
\label{eq:sf_inconsistency}
\end{equation}

\noindent\textbf{L2-Across (L2A).} A measure of pairwise latent-space distance across each step of the candidate path. Here, $Enc$ is an encoder to transform states to a latent space. $\mathcal{H}$ is the set of models ($\langle \hat{P},\hat{R} \rangle \in \mathcal{H}$) learned during meta-training. $\sigma$ is a candidate sequence of actions, representing a possible experimental path. $s_0$ is the starting state from whence $\sigma$ starts, simulating each step forward with $\hat{P}_i$ and $\hat{P}_j$.

\begin{equation}
\text{L2A}(\mathcal{H}, \sigma, s_0) = \sum_{i=1}^{|\mathcal{H}|} \sum_{j=i+1}^{|\mathcal{H}|} \sum_{t=0}^{|\sigma|} \| \text{infer}(s_t, \sigma_t, \hat{P}_i) - \text{infer}(s_t, \sigma_t, \hat{P}_j) \|_2
\label{eq:sf_l2across}
\end{equation}

\noindent\textbf{Central Deviation (CD).} A more computationally feasible approximation of L2-Across. Instead of comparing every model in $\mathcal{H}$, calculate the mean path of latent states induced by $\sigma$, and measure each model's distance from that average.

\begin{equation}
\text{CD}(\mathcal{H}, \sigma, s_0) = \sum_{t = 0}^{|\sigma|} \sum_{\langle\hat{P},\hat{R}\rangle \in \mathcal{H}}
\| \text{infer}(s_t, \sigma_t, \hat{P}) - \mu_t \|_2,
\label{eq:sf_centraldeviation_app}
\end{equation}

Where $\mu_t$ is the average latent vector across all $\hat{P} \in \mathcal{H}$ at timestep t, or:
\[
\mu_t := \frac{1}{|\mathcal{H}|} \sum_{\langle\hat{P},\hat{R}\rangle \in \mathcal{H}} \text{infer}(s_t, \sigma_t, \hat{P})
\]
This function has the desirable property of not requiring a comparison between every pair of models, making it much more feasible to compute.

\subsection{Separating Functions for Stochastic MDPs}
For these, we do not assume the MDPs are deterministic, and so a neural model of $\hat{P}$  returns a distribution over latent state predictions as: 
\[
\text{sample}(s_t, a_t, \hat{P}) := z_{t+1} \sim \hat{P}( \cdot \mid \mathrm{Enc}(s_t), a_t).
\]
Neural dynamics models have a number of methods to represent and sample from these output distributions, but these will not be discussed here. They should all suffice so long as they can be efficiently measured against each other using Kullback–Leibler (KL) divergence.

\noindent\textbf{Pairwise KL (PKL).}  As a stochastic analogue of L2-Across, PKL sums the KL divergence between every pair of model transition distributions at each step, capturing how much models disagree in expectation.  Let $z_t = \mathrm{Enc}(s_t)$ and $a_t = \sigma_t$.  Then

\begin{equation}
\mathrm{PKL}(\mathcal{H},\sigma,s_0)
=\sum_{i=1}^{|\mathcal{H}|}\sum_{j=i+1}^{|\mathcal{H}|}\sum_{t=0}^{|\sigma|}
D_{\mathrm{KL}}\!\bigl(\hat P_i(\cdot\mid z_t,a_t)\,\big\|\;\hat P_j(\cdot\mid z_t,a_t)\bigr).
\label{eq:sf_pairwisekl}
\end{equation}

\noindent\textbf{Central KL Divergence (CKLD).} Extending from central deviation, we measure how much each model’s predicted distribution diverges from the pool’s average prediction at each step, then sum these divergences (using KL divergence).  At step $t$, let $z_t = \mathrm{Enc}(s_t)$ and $a_t = \sigma_t$, and define the mixture
\[
\bar P_t \;=\;\frac{1}{|\mathcal{H}|}\sum_{\langle\hat P,\hat R\rangle\in\mathcal{H}}\hat P\bigl(\cdot\mid z_t,a_t\bigr).
\]
Then the CKLD score is
\begin{equation}
\text{CKLD}(\mathcal{H},\sigma,s_0)
=\sum_{t=0}^{|\sigma|}\sum_{\langle\hat P,\hat R\rangle\in\mathcal{H}}
D_{\mathrm{KL}}\!\bigl(\hat P(\cdot\mid z_t,a_t)\,\big\|\,\bar P_t\bigr).
\label{eq:sf_ckld}
\end{equation}

CKLD quantifies how surprising each model’s prediction is relative to the average prediction—models whose distributions consistently diverge from the mean contribute more KL-mass, signaling that the action sequence produces observations that discriminate between hypotheses. Summing these divergences over time ensures we pick sequences that, on average, maximize the distance between competing models’ beliefs.

Both PKL and CKLD efficiently approximate Mutual Information (MI) — how much observing the outcomes of a planned action sequence tells us about which model in our pool is the true one. PKL approximates MI by summing the KL divergence between every pair of model transition distributions, directly capturing the total discrimination power of a sequence, but it remains more feasible because each divergence can be computed in closed form from the models’ predicted distributions without running a full posterior update (as is required in true MI). CKLD approximates PKL by measuring each model’s KL divergence from the pool’s average prediction, reducing computation from quadratic to linear in the number of models while still rewarding sequences that provoke strong disagreements.

\section{Appendix B: Alchemy Environment}

\label{sec:text_alchemy}

\noindent\textbf{Classical Alchemy Text Conversion.} When converting an underlying state to natural language observations, each element of the state is treated as a binary value, where zero indicates the absence of a specific physical property and one represents its presence. For each property, there are two possible descriptors: one describing its absence (e.g., "smooth") and another describing its presence (e.g., "pointy"). Based on the binary values, one descriptor is selected for each property. Once all the descriptors are chosen, the object being described, such as "stone" or "mineral," is randomly chosen from a pool of generic terms. A descriptive sentence is then created using a pre-designed sentence structure, where placeholders for the object's name and its descriptors are filled in. 

Examples with underlying state and text observation follow:
\begin{itemize}
    \item $[0, 1, 1, 0]$ -- ``The specimen is sharp, shadowy, plain.'' 
    \item $[1, 0, 0, 1]$ -- ``Upon closer inspection, the rock proves to be smooth, dark, colorful.'' 
    \item $[0, 0, 1, 1]$ -- ``Observe this chunk: it appears rounded, shadowy, colorful.'' 
    \item $[1, 1, 0, 0]$ -- ``You notice the mineral is smooth, sharp, plain.'' 
\end{itemize}

\section{Appendix C: Implementation Details}
\label{sec:impl}

\subsubsection{Environments}

HyPE is evaluated on a set of natural language Alchemy environments, variants of the game Alchemy \cite{Wang2021AlchemyAB}. Like the original game, the goal is to transform a stone to a more valuable state before turning it in to earn reward. Stones are transformed by applying potions to them. This will change one of their features, for example, making the stone smooth instead of pointy, or fragrant instead of odorless. In this setting, the underlying state of the stone is represented by a combination of binary traits. These traits form a hyper-cubic graph, as seen in Figure~\ref{fig:cube}. Each potion can only toggle one trait (e.g. moving the state from $[0,0,0]$ to $[0,0,1]$). These potion applications, as well as the single turn-in command that ends the episode, make up the environment's action space. We evaluate on two versions of natural language Alchemy -- a 3D and 4D version. In the 3D version, there are 3 features and 8 states. In the 4D version, there are 4 features and 16 states.

\textbf{Natural Language.} The most clear way natural language Alchemy differs from the original though is its representation of states as the agent sees them. To test HyPE's ability to work with diverse modalities, observations are presented as natural language descriptions of the stone’s current state. For instance, the agent might observe, ``The mineral is sharp, colorful, and dark,'' or ``This sample is smooth and bright, but lacks color.'' Therefore, even if there are only 8 underlying states, there are hundreds of observations the agent may see, requiring generalization and learning of a good latent representation. More details on this language aspect are available in the appendix.

\textbf{Differences in meta-RL tasks.} The transition dynamics and reward function vary across meta-RL worlds. While each potion consistently affects the same trait across worlds, certain actions are blocked from changing the state when applied to specific configurations. For example, in one world, using potion 2 which normally changes the pointiness trait to 1 would be blocked when the agent is at a smooth, colorful, bright stone. A successful agent would then need to take an alternate path. During the meta-training phase, each MDP has 2 blocked transitions in the 3D version, and 4 blocked transitions in the 4D version. These are uniformly sampled for each MDP. 6 MDPs are learned during meta-training for both versions. 

During the adaptation phase, one additional transition is blocked from one of the existing MDPs to create the novel one, simulating environments that are similar to those found in meta-training, but require adaptation nonetheless. This means there is a clear correct environment to select, yet there is still likely adaptation necessary. 40 adaptation trials are conducted, each with a new unseen MDP.

The reward function also varies per MDP. Rewards are assigned based on the presence or absence of specific traits, with each trait contributing differently to the total value. For example, a sharp, dark, and dull stone might have a lower reward, while a bright and colorful stone might yield a higher reward. The maximum reward possible for any state is +1.0, while the minimum is -1.0. Additionally, a small time penalty of -0.05 is applied for each action, encouraging the agent to find efficient paths to maximize rewards.

\subsubsection{Text Encoder}
The text encoder is based on \textit{distilbert-base-uncased} \cite{Sanh2019DistilBERTAD}, a distilled version of the BERT base model \cite{Devlin2019BERTPO}. It was chosen for its smaller size and faster speed, making it less prone to overfitting while retaining strong pre-trained sentence-level embeddings. The model has an output latent space of 768 dimensions, which is further reduced with a linear layer to 64 units. Language models like this produce a fixed-size vector by first breaking the text into sub-word units. These tokens are converted into embedding vectors from a learned table, augmented with positional encodings to capture word order. The embeddings pass through many transformer layers, where self-attention mechanisms encode relationships across the entire input, refining token representation in each layer. The output embedding of the special [CLS] token provides a good fixed-size vector summarizing the input for downstream tasks.

\subsubsection{Latent Dynamics Model}

The latent dynamics model is a two-layer feedforward network. The input consists of the current latent state vector concatenated with an action vector. The hidden layers have 256 and 32 units, respectively. The model outputs the predicted next latent state, the predicted reward, and a binary terminal indicator to determine whether the episode will end. 

Internally, it operates as a "delta model", predicting the change in the current latent state, which is added to the input vector to compute the next latent state. This is not strictly necessary to our work, but it does make learning the dynamics much easier empirically, and it is a well-known technique with its roots in control theory.

\subsubsection{Latent Actor}

The latent actor is a Model-Predictive Controller (MPC) agent that plans actions using the learned latent dynamics model. It evaluates sequences of actions by predicting their cumulative reward and whether the episode will terminate. Using a random shooting search, the actor explores 20,000 possible action sequences within a fixed horizon (in this case, 5 steps ahead), selecting the one with the highest predicted reward. After each action, the plan is re-calculated using the updated state to minimize error accumulation from the model.

\subsubsection{HyPE Settings}
Central deviation is selected as a separation function, and a random shooting search similar to that used by the actor was used to generate candidate experimental action sequences. The FAMLE algorithm \cite{kaushik2020fast} was followed for meta-training and adaptation outside of model selection -- HyPE being a drop-in replacement for the standard passive exploration here. See more implementation details in the appendix.

\subsubsection{Passive Exploration Baseline}
We compare HyPE with a standard passive exploration strategy -- explore-then-commit (EtC). Simply, random action selection is undertaken for $k$ steps, then the model is chosen based on what best fits the randomly-collected experience. The same FAMLE algorithm as HyPE implemented all other meta-RL components. This baseline only differed from HyPE in the key model selection part of the adaptation phase. See Figure~\ref{fig:algo} left side.

\textbf{Meta-training Phase.} The meta-training phase used the Adam optimizer with a learning rate of 0.00005 and a batch size of 512. Six tasks were learned, with each task generating 25,600 random training transitions for offline model learning. Additionally, 512 transitions per task were set aside as a small validation set, also utilized for early stopping, though this condition was never triggered. Training ran for 1,000 epochs, during which the encoder was frozen after 400 epochs to allow the latent dynamics models to stabilize within the latent space. A binary feature classifier, consisting of a two-layer neural network with 128 hidden units, was employed to pretrain the encoder by predicting the underlying state from the latent space. The pool of trained models generated during this phase was used for both HyPE and the passive exploration baseline.

\textbf{Adaptation Phase.} The adaptation phase employed an SGD optimizer with a learning rate of 0.00001 and a batch size of 16. The model was updated online once per epoch using data from an experience buffer, applying the same procedure for both HyPE and the passive exploration baseline.

\textbf{Computational Details.} All experiments were conducted on an NVIDIA A100 GPU. The meta-training process required 8–12 hours to complete 1,000 epochs. The adaptation phase required roughly 2 minutes per trial.
\section{Appendix D: Proofs}

\subsection{Proof of Theorem \ref{thm:ior}}
\label{sec:proof_thm_ior}

Here, we discuss and then prove Theorem \ref{thm:ior} from the main body of the paper.

Formally, define $\mathcal G$ as:
\[
\mathcal G := \left\{(s,a)\in\mathcal S \times\mathcal A : 
\operatorname{div}(s,a) \ge d_0 \right\},
\]
\[
\operatorname{div}(s,a) := \min_{i\ne j} D_{\mathrm{KL}}(P_i(\cdot\mid s,a)\|P_j(\cdot\mid s,a)).
\]

Here, $P_i(\cdot\mid s,a)$ denotes the next-state distribution in MDP $i$ given state-action pair $(s,a)$, and $D_{\mathrm{KL}}(\cdot|\cdot)$ is the Kullback–Leibler divergence between two such distributions. The threshold $d_0 > 0$ determines when the divergence is considered meaningfully informative for distinguishing MDPs. 
Intuitively, $\mathcal G$ consists of the space of transitions that yield substantial information about which MDP is correct, where ``substantial'' is quantified by some lower bound $d_0 > 0$.

For any policy \(\pi\), define its time‑averaged occupancy measure,
\[
\mu_{\pi}(s,a)
\;=\;
\frac{1}{T}\sum_{t=1}^T
\Pr_{\tau\sim P_i^{\pi}}\bigl((s_t,a_t)=(s,a)\bigr),
\]
i.e.\ the fraction of the \(T\) steps spent at \((s,a)\).  Then the fraction of visits to informative transitions under \(\pi\) is
\[
\varepsilon
\;=\;
\sum_{(s,a)\in\mathcal G}\mu_{\pi_{uni}}(s,a),
\quad
\alpha
\;=\;
\sum_{(s,a)\in\mathcal G}\mu_{\pi_{hype}}(s,a),
\]
Small \(\varepsilon\) indicates that uniform‐action exploration rarely encounters \(\mathcal G\), whereas a larger \(\alpha\) shows that planned exploration effectively targets it.

To prove the theorem, we introduce the following Lemma.
\begin{lemma}[Policy‐trajectory upper bound]
\label{lem:policy}
Let $\pi$ be any fixed policy mapping states to action distributions over $\mathcal A$, and let
\[
\Delta_{\pi}
\;=\;
\min_{i \neq j}
D_{\mathrm{KL}}\!\bigl(P_i^{\pi}(\tau)\,\big\|\,P_j^{\pi}(\tau)\bigr),
\]
where $P_i^{\pi}(\tau)$ is the distribution over $T$‑step trajectories under MDP $i$ and policy $\pi$.  Then
\[
P_{\pi}(\hat i \neq i^*) \;\le\; (|\mathcal{H}|-1)\,\exp\bigl(-\Delta_{\pi}\bigr).
\]
\end{lemma}
\noindent This is proven in Section \ref{lem:policy}. Then we apply this lemma to the uniform and planned policies to characterize their relative sample efficiencies when $\alpha>\varepsilon$.

\begin{proof}
\noindent\textbf{Remark.} By the chain rule for KL divergence—and since all MDPs share the same initial distribution \(\rho\)—we have
\[
D_{\mathrm{KL}}\bigl(P_i^\pi(\tau)\big\|P_j^\pi(\tau)\bigr)
=\mathbb{E}_{\tau\sim P_i^\pi}\!\biggl[\sum_{t=1}^T
D_{\mathrm{KL}}\bigl(P_i(\cdot\mid s_t,a_t)\big\|P_j(\cdot\mid s_t,a_t)\bigr)\biggr].
\]

\noindent\textbf{Step 1. Uniform policy bound.}  
Apply with $\pi=\pi_{uni}$ to get
\[
P_{\pi_{uni}}(\hat i \neq i^*)
\;\le\;
(|\mathcal{H}|-1)\,\exp\bigl(-\Delta_{\pi_{uni}}\bigr).
\]
\noindent\textbf{Intuition.}  Under uniform random actions, each trajectory gives information distinguishing the true MDP \(i^*\) from any other candidate MDP \(j\) (the \textit{competitors}).  The hardest competitor is the one whose trajectory distribution is closest (i.e.\ has the smallest KL divergence) to that of \(i^*\); that minimal gap is exactly \(\Delta_{\pi_{uni}}\).  Hence, the probability of confusing \(i^*\) with any competitor decays exponentially at rate \(\Delta_{\pi_{uni}}\), up to the \((|\mathcal{H}|-1)\) possible wrong choices.

\noindent\textbf{Step 2. Relate $\Delta_{\pi_{uni}}$ to $\varepsilon\,d_0\,T$.}  
By definition,
\[
\Delta_{\pi_{uni}}
=\min_{i\neq j}\;
\mathbb{E}_{\tau\sim P_i^{\pi_{uni}}}\Bigl[\sum_{t=1}^T
D_{\mathrm{KL}}\bigl(P_i(\cdot\mid s_t,a_t)\,\|\,P_j(\cdot\mid s_t,a_t)\bigr)\Bigr].
\]
Since $D_{\mathrm{KL}}\ge d_0$ on $\mathcal G$ and $\ge0$ elsewhere,
\[
\Delta_{\pi_{uni}}
\;\ge\;
d_0\sum_{t=1}^T\Pr\bigl((s_t,a_t)\in\mathcal G\bigr)
\;=\;
d_0\,(\varepsilon\,T).
\]
Hence
\[
P_{\pi_{uni}}(\hat i \neq i^*)
\;\le\;
(|\mathcal{H}|-1)\,\exp\bigl(-\varepsilon\,d_0\,T\bigr).
\]
\textbf{Intuition.}  Under uniform‐action, informative transitions occur only an $\varepsilon$ fraction of the time.

\textbf{Remark.} Even if $(s,a) \notin \mathcal{G}$, transitions may still yield small divergence $\bar{d} > 0$. These accumulate information slowly, requiring approximately $1/\bar{d}$ more steps to match the confidence achieved by transitions in $\mathcal{G}$, so excluding them from the bound does not change its qualitative behavior when $\bar{d} \ll d_0$.

Let
\[
\bar{d} := \max_{(s,a) \notin \mathcal{G}} \min_{i \ne j} D_{\mathrm{KL}}\big(P_i(\cdot \mid s,a) \,\|\, P_j(\cdot \mid s,a)\big)
\]
denote the greatest divergence between any two models over transitions outside $\mathcal{G}$. $\bar{d}$ measures the strongest available -- but still sub-threshold—distinguishing signal from weak transitions. If the agent never visits $\mathcal{G}$, $\bar{d}$ bounds the per-step rate of information gain, and total confidence can only grow at this slower rate.

\noindent\textbf{Step 3. Planned policy bound.}  
Apply Lemma~\ref{lem:policy} with $\pi=\pi_{hype}$:
\[
P_{\pi_{hype}}(\hat i \neq i^*)
\;\le\;
(|\mathcal{H}|-1)\,\exp\bigl(-\Delta_{\pi_{hype}}\bigr).
\]
\textbf{Intuition.}  The planned policy yields a higher minimal trajectory‐KL, $\Delta_{\pi_{hype}}$.

\noindent\textbf{Step 4. Relate $\Delta_{\pi_{hype}}$ to $\alpha\,d_0\,T$.}  
By the same occupancy argument,
\[
\Delta_{\pi_{hype}}
\;\ge\;
d_0\sum_{t=1}^T\Pr\bigl((s_t,a_t)\in\mathcal G\bigr)
\;=\;
d_0\,(\alpha\,T).
\]
Thus
\[
P_{\pi_{hype}}(\hat i \neq i^*)
\;\le\;
(|\mathcal{H}|-1)\,\exp\bigl(-\alpha\,d_0\,T\bigr).
\]
\textbf{Intuition.}  Planned exploration visits informative pairs an $\alpha$ fraction of the time.

\noindent\textbf{Step 5. Sample‐complexity ratio.}  
Fix target error $\delta\in(0,1)$.  Solving
\[
(|\mathcal{H}|-1)\exp\bigl(-\varepsilon\,d_0\,T_{\mathrm{uni}}\bigr)\le\delta
\quad\Longrightarrow\quad
T_{\mathrm{uni}}\ge\frac{\ln\!\bigl((|\mathcal{H}|-1)/\delta\bigr)}{\varepsilon\,d_0},
\]
and
\[
(|\mathcal{H}|-1)\exp\bigl(-\alpha\,d_0\,T_{\mathrm{hype}}\bigr)\le\delta
\quad\Longrightarrow\quad
T_{\mathrm{hype}}\ge\frac{\ln\!\bigl((|\mathcal{H}|-1)/\delta\bigr)}{\alpha\,d_0},
\]
gives
\[
\frac{T_{\mathrm{hype}}}{T_{\mathrm{uni}}}
\;\le\;
\frac{\varepsilon}{\alpha}
\;=\;\frac1{\mathrm{IOR}}.
\]
\textbf{Intuition.}  Planned exploration needs only a $1/\mathrm{IOR}$ fraction of the samples.

%\noindent\textbf{Step 6. Error‐ratio at fixed $T$.}  
%Dividing the two bounds yields
%\[
%\frac{P_{\pi_{uni}}(\hat i\neq i^*)}
%     {P_{\pi_{hype}}(\hat i\neq i^*)}
%\;\le\;
%\exp\!\bigl((\alpha-\varepsilon)\,d_0\,T\bigr).
%\]
%\textbf{Intuition.}  Whenever $\alpha>\varepsilon$, planned exploration achieves exponentially smaller error.

\noindent\textbf{Step 6. Error‐ratio at fixed $T$.}  
From Steps 2 and 4 we have
\[
P_{\pi_{uni}}(\hat i \neq i^*)
\;\le\;
(|\mathcal{H}|-1)\,\exp\bigl(-\varepsilon\,d_0\,T\bigr),
\quad
P_{\pi_{hype}}(\hat i \neq i^*)
\;\le\;
(|\mathcal{H}|-1)\,\exp\bigl(-\alpha\,d_0\,T\bigr).
\]
Dividing the second by the first (the \((|\mathcal{H}|-1)\) cancels) yields
\[
\frac{P_{\pi_{hype}}(\hat i \neq i^*)}
     {P_{\pi_{uni}}(\hat i \neq i^*)}
\;\le\;
\frac{\exp(-\alpha\,d_0\,T)}{\exp(-\varepsilon\,d_0\,T)}
\;=\;
\exp\!\bigl(-(\alpha - \varepsilon)\,d_0\,T\bigr).
\]

\noindent\textbf{This completes the proof of Theorem \ref{thm:ior}.}
\end{proof}

\newpage
\subsection{Proof of Lemma \ref{lem:policy}}
\label{sec:proof_lem_pol}
\begin{proof}
Recall that under each MDP $i$ and a fixed policy $\pi$, a $T$‑step trajectory
\[
\tau=(s_1,a_1,\dots,s_T,a_T,s_{T+1})
\]
has probability
\[
P_i^{\pi}(\tau)
=\rho(s_1)\prod_{t=1}^T\pi(a_t\mid s_t)\,P_i(s_{t+1}\mid s_t,a_t),
\]
where $\rho$ is the common initial‐state distribution (which cancels out in any likelihood ratio).

\noindent\textbf{Step 1.}  For each competitor $j\neq i^*$, define the trajectory log‑likelihood ratio
\[
Z^{(j)}(\tau)
:= \ln \frac{P_{i^*}^{\pi}(\tau)}{P_{j}^{\pi}(\tau)}.
\]

\noindent\textbf{Intuition.}  $Z^{(j)}(\tau)$ quantifies how much more likely the observed trajectory is under the true MDP $i^*$ than under the wrong model $j$.

\noindent\textbf{Step 2.}  The event $\{\hat i \neq i^*\}$ implies $\exists\,j\neq i^*$ with $Z^{(j)}(\tau)\le 0$.  Hence by a union bound,
\[
P_{\pi}(\hat i \neq i^*)
= P\bigl(\exists\,j\neq i^*:Z^{(j)}\le0\bigr)
\;\le\;
\sum_{j\neq i^*}P\bigl(Z^{(j)}\le0\bigr).
\]

\noindent\textbf{Intuition.}  Misidentification happens if any wrong model explains the data at least as well as the true one.

\noindent\textbf{Step 3.}  For fixed $j$ and any $0<\lambda<1$,
\[
P\bigl(Z^{(j)}\le0\bigr)
= P\bigl(e^{-\lambda Z^{(j)}}\ge1\bigr)
\;\le\;
\mathbb{E}\bigl[e^{-\lambda Z^{(j)}}\bigr]
=\sum_{\tau}
P_{i^*}^{\pi}(\tau)^{1-\lambda}\,
P_{j}^{\pi}(\tau)^{\lambda}.
\]

\noindent\textbf{Intuition.}  Markov’s inequality on $e^{-\lambda Z^{(j)}}$ converts the tail event into an expectation involving both trajectory distributions.

\noindent\textbf{Step 4.}  Optimizing over $\lambda$ yields a KL-based Exponential Tail Bound via Markov’s Inequality.
\[
P\bigl(Z^{(j)}\le0\bigr)
\;\le\;
\exp\!\bigl(-D_{\mathrm{KL}}\bigl(P_{i^*}^{\pi}\,\|\,P_{j}^{\pi}\bigr)\bigr).
\]

\noindent\textbf{Intuition.}  The best exponential decay rate is exactly the KL divergence between the two trajectory distributions.

\noindent\textbf{Step 5.}  Applying the union bound across all $j\neq i^*$,
\[
P_{\pi}(\hat i \neq i^*)
\;\le\;
\sum_{j\neq i^*}\exp\!\bigl(-D_{\mathrm{KL}}(P_{i^*}^{\pi}\|P_{j}^{\pi})\bigr)
\;\le\;
(|\mathcal{H}|-1)\,\exp\bigl(-\Delta_{\pi}\bigr),
\]
where
\[
\Delta_{\pi}
=\min_{i\neq j}
D_{\mathrm{KL}}\!\bigl(P_i^{\pi}(\tau)\,\big\|\,P_j^{\pi}(\tau)\bigr).
\]

\noindent\textbf{Intuition.}  The smallest separating KL divergence, amplified by the union bound, yields the final exponential error bound.

\noindent\textbf{This completes the proof of Lemma \ref{lem:policy}.}
\end{proof}

\subsection{Theorem on Finding the Closest Hypothesis}

Further, we can maintain this efficiency when we no longer assume that the true MDP we are interacting with is in the hypothesis pool (as is expected for meta-RL). The task then becomes finding the closest model in the pool.

\begin{theorem}
\label{thm:closest}

Under the same definitions of $\alpha,\varepsilon,\mathrm{IOR}$ as in Theorem~\ref{thm:ior}, let $c$ be the reference model whose transitions are, in the worst case, closest (in KL divergence) to those of the true MDP. Define $\gamma>0$ to be the smallest per–state–action KL advantage that $M_c$ has over any other reference model.  Then for any horizon $T$,
\[
\frac{P_{\pi_{\text{hype}}}(\hat i \neq c)}
     {P_{\pi_{\text{uni}}}(\hat i \neq c)}
\;\le\;
\exp\!\bigl(-(\alpha - \varepsilon)\,\gamma\,T\bigr).
\]
\end{theorem}

\newpage
\section{Proof of Theorem \ref{thm:closest}}
\begin{proof}
We follow the same structure as in Theorem~\ref{thm:ior}, replacing \(d_0\) by \(\gamma\) and \(i^*\) by \(c\).

\noindent\textbf{Step 1. Apply the policy‐trajectory bound.}  
By Lemma~\ref{lem:policy}, for any fixed policy \(\pi\),
\[
P_{\pi}(\hat i \neq c)
\;\le\;
(|\mathcal{H}|-1)\,\exp\bigl(-\Delta_{\pi}\bigr),
\]
where
\[
\Delta_{\pi}
=\min_{j\neq c}\Bigl\{
D_{\mathrm{KL}}\!\bigl(P_{k+1}^{\pi}\|\;P_j^{\pi}\bigr)
-\;D_{\mathrm{KL}}\!\bigl(P_{k+1}^{\pi}\|\;P_c^{\pi}\bigr)
\Bigr\}.
\]
\textbf{Intuition.}  This captures the worst‐case average log‐likelihood gap favoring \(c\) over any other model.

\noindent\textbf{Step 2. Lower‐bound \(\Delta_{\pi_{\mathrm{uni}}}\).}  
Under the uniform policy,
\[
\Delta_{\pi_{\mathrm{uni}}}
\;=\;\min_{j\neq c}\;\mathbb{E}\Bigl[\sum_{t=1}^T
\bigl(d_{k+1,j}(s_t,a_t)-d_{k+1,c}(s_t,a_t)\bigr)\Bigr],
\]
where \(d_{i,j}(s,a)=D_{\mathrm{KL}}(P_i\|\!P_j)(s,a)\).  By definition of \(\gamma\), each term in \(\mathcal G\) contributes at least \(\gamma\), and zero elsewhere.  Hence
\[
\Delta_{\pi_{\mathrm{uni}}}
\;\ge\;
\gamma \sum_{t=1}^T \Pr\bigl((s_t,a_t)\in\mathcal G\bigr)
=\gamma\,(\varepsilon\,T).
\]
\textbf{Intuition.}  Uniform exploration only accrues evidence when it accidentally hits a distinguishing transition.

\noindent\textbf{Step 3. Lower‐bound \(\Delta_{\pi_{\mathrm{hype}}}\).}  
Similarly, for the planned policy,
\[
\Delta_{\pi_{\mathrm{hype}}}
\;\ge\;
\gamma \sum_{t=1}^T \Pr\bigl((s_t,a_t)\in\mathcal G\bigr)
=\gamma\,(\alpha\,T).
\]
\textbf{Intuition.}  HyPE deliberately seeks out those transitions, accruing \(\gamma\) whenever it does.

\noindent\textbf{Step 4. Error‐ratio at fixed $T$.}  
From the bounds
\[
P_{\pi_{hype}}(\hat i \neq c)\le(K-1)\exp(-\alpha\,\gamma\,T),
\quad
P_{\pi_{uni}}(\hat i \neq c)\le(K-1)\exp(-\varepsilon\,\gamma\,T),
\]
divide the first by the second (cancelling \((K-1)\)) to get
\[
\frac{P_{\pi_{\text{hype}}}(\hat i \neq c)}
     {P_{\pi_{\text{uni}}}(\hat i \neq c)}
\;\le\;
\frac{\exp(-\alpha\,\gamma\,T)}{\exp(-\varepsilon\,\gamma\,T)}
=\exp\!\bigl(-(\alpha - \varepsilon)\,\gamma\,T\bigr).
\]
\textbf{Intuition.}  Since HyPE visits informative transitions an $\alpha$ fraction of the time (vs.\ $\varepsilon$ for uniform), its trajectory‐KL grows faster by $(\alpha-\varepsilon)\gamma\,T$, yielding an exponentially smaller error ratio.

\noindent\textbf{This completes the proof of Theorem \ref{thm:closest}.}
\end{proof}


\end{document}